\documentclass[sigconf]{aamas}  

\usepackage{booktabs}

\setcopyright{ifaamas}  
\acmDOI{doi}  
\acmISBN{}  
\acmConference[AAMAS'19]{Proc.\@ of the 18th International Conference on Autonomous Agents and Multiagent Systems (AAMAS 2019), N.~Agmon, M.~E.~Taylor, E.~Elkind, M.~Veloso (eds.)}{May 2019}{Montreal, Canada}  
\acmYear{2019}  
\copyrightyear{2019}  
\acmPrice{}  

\begin{document}

\title{MaMiC: Macro and Micro Curriculum for Robotic Reinforcement Learning}  




%

\author{Manan Tomar$^{1}$, Akhil Sathuluri$^{1}$, Balaraman Ravindran$^{1,2}$}
\affiliation{%
  \institution{$^{1}$Indian Institute of Technology Madras, Chennai, India \\ $^{2}$Robert Bosch Center for Data Science and AI (RBCDSAI), Chennai, India}
  {manan.tomar@gmail.com,  akhilsathuluri@gmail.com,  ravi@cse.iitm.ac.in}
}

\begin{abstract}  

Shaping in humans and animals has been shown to be a powerful tool for learning complex tasks as compared to learning in a randomized fashion. This makes the problem less complex and enables one to solve the easier sub task at hand first. Generating a curriculum for such guided learning involves subjecting the agent to easier goals first, and then gradually increasing their difficulty. This paper takes a similar direction and proposes a dual curriculum scheme for solving robotic manipulation tasks with sparse rewards, called MaMiC. It includes a macro curriculum scheme which divides the task into multiple sub-tasks followed by a micro curriculum scheme which enables the agent to learn between such discovered sub-tasks. We show how combining macro and micro curriculum strategies help in overcoming major exploratory constraints considered in robot manipulation tasks without having to engineer any complex rewards. We also illustrate the meaning of the individual curricula and how they can be used independently based on the task. The performance of such a dual curriculum scheme is analyzed on the Fetch environments. 

\end{abstract}

%


\maketitle



\section{Introduction}
In recent years, deep Reinforcement Learning has seen a lot of promising results in varied domains such as game-playing \cite{mnih2015human}, \cite{silver2016mastering}, and continuous control \cite{lillicrap2016continuous}, \cite{schulman2015trust}, \cite{gu2017deep}. Despite these developments, robotic decision making remains a hard problem given minimal context of the task in hand \cite{deisenroth2013survey}. Robotic learning presents a huge challenge mainly because of the complex dynamics, sparse rewards and exploration issues arising from large continuous state spaces, thus providing a good testbed for reinforcement learning algorithms.

Solving complex tasks requires exploiting the structure of the task efficiently. 
Each task can be viewed as a combination of much simpler prelearnt skills. Consider the cases of unscrewing a bottle or placing an object in a drawer. All such everyday tasks involve reusing distinct skills or sub-policies in an intelligent manner to achieve the overall objective. To be able to solve such complex tasks it is important that we learn in a organized, meaningful manner rather than learning using data collected in a random fashion. Curriculum learning \cite{bengio2009curriculum}, \cite{sukhbaatar2017intrinsic} is a powerful concept that allows us to come up with such training strategies. Starting to learn for simpler tasks and then using the acquired knowledge to learn progressively harder tasks is a natural outcome of formulating a curriculum. A curriculum assists one in overcoming exploratory constraints of the agent by focusing learning over simpler parts of the state space first.
Recently, curriculum learning has been used to solve complex robotic tasks (not necessarily manipulation) such as in \cite{florensa2017reverse}, \cite{nair2017overcoming}.
However, these approaches make the assumption that the agent can be reset to any desired state, and also make use of expert state action trajectories \cite{nair2017overcoming}, which are expensive to generate. Unlike such techniques, our method is not restricted by the ability to reset. Moreover, we use state-only demonstration sequences for learning only in specific tasks, and do not use demonstrations at all for the other tasks, thus distinguishing our work from those in the imitation learning sphere. Although learning from only state or observation sequences is a much more difficult method \cite{liu2018imitation}, it offers practical benefits in terms of reduced trajectory collection costs and implementation ease, thus fitting our problem domain more accurately.

One way of looking at the problem in hand is to extract sub-goals for a given task, learn sub-policies or skills that achieve these sub-goals, and then execute them in the right order. Such a top-down approach allows exploiting the structure of the problem, since the extracted sub-goals define the nature of the solution. Moreover, we also focus on the sequential nature of the problem, i.e. solving to achieve the first sub-goal, then the second sub-goal and so on. This is important as most robotic locomotion or manipulation problems can be recognized in this manner. In our method, the sub-goal extraction and sequencing is managed by the macro scheme, while learning each sub-policy is managed by the micro scheme. In order to achieve this, both of these methods exhibit and use concepts from curriculum learning. 

\begin{figure*}[h]
	\centering
	\includegraphics[width=0.85\textwidth]{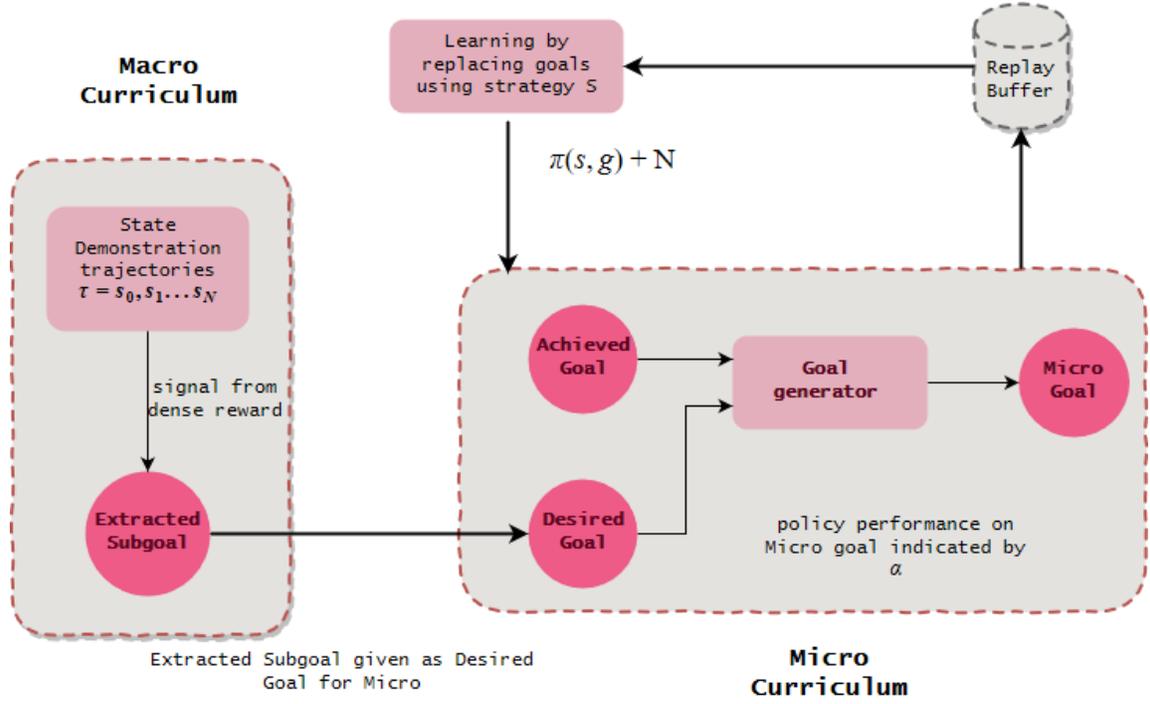}
    \caption{A diagram illustrating the working of MaMiC }
    \label{diagram}
\end{figure*}

We introduce MaMiC, comprising macro and micro curriculum, which can be applied either individually or in combination. A micro curriculum essentially generates increasingly complex goals for the agent to achieve. For example, in learning to push a block, initial goals will be generated very near to the block and then slowly shifted to the desired location. However, such a scheme is not sufficient if we need to solve tasks which are more complex, such as ones which require the agent to maintain a particular sequence of sub policies. For instance, in order to put an object in a drawer, it is not enough to guide the agent in learning to put the object to the desired location, but also to open the drawer first. It is only when a particular sequence of such sub policies is followed that we refer to the task as completed. A macro curriculum helps in identifying such a sequence and allows the micro scheme to learn in between this sequence. A policy starts from a achieved sub-goal and proceeds to the next sub-goal, evolving in the process, ultimately reaching the actual goal. Two ideas are at the core of this technique, of being able to discover the sub-goals and of learning between the recognized sub-goals. The working of MaMiC is as illustrated in Fig \ref{diagram}. To summarize, the following are the major contributions of the paper:





\begin{itemize}
\item We propose a dual curriculum strategy comprising micro and macro schemes, which enables an agent to discover sub-goals and learn a policy which evolves to achieve such sub-goals sequentially, eventually solving the task
\item We analyze the macro and micro schemes individually, and illustrate how to combine these individual schemes with base reinforcement learning algorithms such as Deep Deterministic Policy Gradients (DDPG) to solve a given task
\item The performance of the proposed dual curriculum scheme is tested in a Receptor-PickandPlace environment and also in a custom physics environment. 
\item An industrial robot with minimal observations available is considered for training and the learnt policy is deployed onto a physical robot as validation. (see the supplementary videos at https://goo.gl/nKZoCQ)
\end{itemize}

\section{Background and Preliminaries}
Reinforcement Learning (RL), \cite{sutton1998reinforcement}, considers the interaction of an agent with a given environment and is modeled by a Markov Decision Process (MDP), defined by the tuple $\mathcal{\langle S, A, P, \rho_{0}}, r \rangle$, where $\mathcal{S}$ defines the set of states, $\mathcal{A}$ the set of actions, $\mathcal{P  : S \times A \rightarrow S}$ the transition function, $\mathcal{\rho_{0}}$ the probability distribution over initial states, and $r\mathcal{ : S \times A \rightarrow R}$ the reward function. A policy is denoted by $\pi(s) \mathcal{: S \rightarrow P(A)} $, where $\mathcal{P(A)}$ defines a probability distribution over actions $a \mathcal{ \ \epsilon \ A}$ in a state $s \mathcal{ \ \epsilon \ S}$. The objective is to learn a policy such that the return $R_{t} = \sum_{i=t}^{T} \ \gamma^{(i-t)} \ r(s_{i}, a_{i})$ is maximized, where $r(s_{i}, a_{i}) \ $ is the reward function and $\gamma$ is the discount factor.

\subsection{Deep Deterministic Policy Gradients (DDPG)}
DDPG \cite{lillicrap2016continuous} is an off policy, model free actor critic based reinforcement learning method. The critic is used to estimate the action value function $Q(s_{t}, a_{t})$, while the actor refers to the deterministic policy of the agent. The critic is learned by minimizing the standard TD error


\begin{equation}
\delta_{TD} = r_{t} + \gamma Q^{\prime}(s_{t+1}, \pi(s_{t+1})) \ -  \ Q(s_{t}, a_{t})
\end{equation}

\vspace{2mm}
,where $Q^{\prime}$ refers to a target network \cite{mnih2015human} which is updated after a fixed number of time steps. The actor is optimized by following the gradient of the critic's estimate of the $Q$ value.
Universal Value Function Approximators (UVFA) \cite{schaul2015universal} parameterizes the $Q$ value function by the goal and tries to learn a policy $\pi(s_{t}, g_{t}) \mathcal{: S \times G }\rightarrow A$ dependent on the goal as well. Such a value function is denoted by $Q(s, a, g)$. 

\subsection{Hindsight Experience Replay}
Hindsight Experience Replay (HER) was introduced by \cite{andrychowicz2017hindsight} and works along with an off policy method such as DDPG to accelerate the learning process. The overall idea is to learn from unsuccessful trials as well by parameterizing over goals. HER helps in accelerating learning by substituting some samples with the \texttt{{achieved goal}} instead of the actual goal. Since the current policy is able to reach these \texttt{{achieved goal}}, learning the mapping between goals to actions becomes faster.

\subsection{Goal Generative Adversarial Network}
\cite{held2017automatic} propose using a Generative Adversarial Network (GAN) \cite{goodfellow2014generative}, \cite{mao2017least} based goal generator for sampling good goals, which refers to goals which are neither too hard nor too easy for the current policy to achieve. The goals used for training the GAN are labeled based on the return obtained for the specific goal. Goals which lead to a positive return are encouraged while those which lead to a negative return are discouraged.


\subsection{Definitions}
The following are the definitions of the terms used throughout the paper:

\texttt{\textbf{Desired Goals :}}
These refer to the actual goals received from the environment and correspond to the task being solved.

\texttt{\textbf{Achieved Goals :}}
These refer to end of trajectory states achieved by the agent while following the currently learned policy.

\texttt{\textbf{HER Goals:}}
These refer to achieved states in a trajectory while following the currently learned policy, randomly sampled as is proposed by HER .

\texttt{\textbf{Micro Goals:}}
These refer to goals generated by the goal generator.

\texttt{\textbf{Sub Goals:}}
These refer to the \texttt{sub-goals} extracted from demonstrations or assumed to be given by an oracle.

This work assumes that there exists a mapping $m(g) : G \rightarrow S \ $ between a goal $g \ \epsilon \ G$ and a state $s \ \epsilon \ S$. The task then is defined by achieving the corresponding goal state $s_{g}$ for a given goal $g$. Note that if such a mapping exists, a goal can be achieved by achieving more than one state. Many robotic manipulation tasks are designed such that the goal can be represented as an achievable state, and therefore, such an assumption does not add extreme constraints. In such cases, the achieved goal can be the object's position and the desired goal can be the target location. Note that the framework adopted in this work does not limit us to only have Cartesian coordinates of objects for defining an achieved goal. 

Assumptions about the environment dilute the generalization of an algorithm and lead to failure in unconstrained settings or real-world deployment. In manipulation tasks these can be alleviated by breaking the task into much simpler tasks with lesser constraints, making it easier for the agent to learn. Below are two such assumptions:

\begin{itemize}
	\item \textbf{Resetting the agent to any desired state :}
	In the native reinforcement learning setting, the agent is initialized at particular states based on a start state distribution available only to the environment. However, as mentioned, previous works assume that the agent can be initialized from whichever state is desired. Given this assumption, the agent can start directly from the goal state and thus not learn at all. Such an assumption is extremely limiting as in any practical setting the environment dictates the start state of the agent. The agent should be intelligent enough to reach such desired or favorable states. 
    
	\item \textbf{Starting from solved or partially solved states :}
	Prior work also mentions another technique for learning sparse reward manipulation tasks which involves starting some training trajectories from solved states and the rest by sampling from the start state distribution. For example, in a pushing task, some trajectories start with the object being placed at the target location.
\end{itemize}

\section{Micro Curriculum}
\vspace{2mm}
A micro curriculum tries to alleviate the above-mentioned assumption of being able to start some trajectories from favorable states. As argued above, we believe that starting at a particular state should be based on the environment's choice but not the agent's. We propose replacing all or some transition sample goals with the \texttt{micro goals} which may be generated by any generative modeling technique. Using an off policy RL algorithm allows us to replace sampled transition goals from the buffer with \texttt{micro goals}. The goals are generated such that they are initially close to the achieved states at the end of each trajectory (i.e. the \texttt{{achieved goal}} distribution) and slowly shift to being closer to the actual or \texttt{{desired goal}} distribution of the task in hand. Since this procedure involves learning a mapping between goals and actions, eventually the agent is able to generalize well for the actual goal distribution. We relate this with curriculum learning because the agent initially learns for a goal distribution much simpler to learn i.e. the  \texttt{{achieved goal}} distribution and then continues learning for increasingly difficult goals, leveraging the previously learned skills.

\vspace{2mm}
To train the goal generator, we make use of Generative Adversarial Networks or GANs \cite{goodfellow2014generative} and modify the formulation used by \cite{held2017automatic}. We incorporate an additional parameter $\alpha \ \epsilon \ [0, 1]$ which governs the resemblance of the generated distribution to the \texttt{{achieved goal}} distribution and the actual or \texttt{{desired goal}} distribution. $\alpha = 0$ forces the generator to produce goals similar to the currently achieved states, while $\alpha = 1$ produces goals similar to the actual distribution. The exact objective function is given below.

\begin{eqnarray}\nonumber
\small{min_{D} V(D)} = \small{\mathbf{E}_{g \sim p_{data}(g)}[(1 - \alpha) \ (D(g_{achieved}) - 1)^{2} + } \\
\small{\ \alpha \ (D(g_{desired}) - 1)^{2}] \ + \ \mathbf{E}_{z \sim p_{z}(z)}[D(G(z))^{2}]}
\end{eqnarray}

\begin{equation}
\small{min_{G} \ V(G) \ = \mathbf{E}_{z \sim p_{z}(z)}[(D(G(z)) \ - 1)^{2}]}
\end{equation}

\vspace{2mm}

\hspace{2mm},where $D$ denotes the discriminator network, $G$ the generator network, and $V$ the GAN value function. $p_z$ here is taken as a uniform distribution between 0 and 1 from which the noise vector $z$ is sampled. In all experiments that follow, we choose to update $\alpha$ if the success rate of the currently learned policy for goals generated by the GAN lies above a particular threshold consistently for a few epochs. This essentially tells us that the policy has now mastered achieving the currently generated goals with some degree of confidence and thus the GAN can now  shift further towards producing goals resembling the desired distribution.


\noindent\rule[1.4pt]{0.5\textwidth}{1.0pt}
\vspace{-2mm}
\noindent \textbf{Algorithm 1 :} Micro Curriculum

\noindent\rule[1.4pt]{0.5\textwidth}{1.0pt}
\setlength{\parskip}{0.0ex}

\textbf{Given :} An off policy RL algorithm $A$, a goal generator $G$, a goal sampling strategy $S$, replay buffer $R$ 

\vspace{2mm}

\noindent Initialize $A$, $R$, $G$


\noindent $\textbf{for} \ n = 1, ..., N$ episodes \textbf{do} 

\hspace{4mm} Sample initial state $s_{0}$, goal $g$ 

\hspace{4mm} $\texttt{DesiredGoal}_{n} \leftarrow g$

\hspace{4mm} Generate artificial goal from $G, \ g_{micro} \leftarrow G(z)$, $z \sim p_z$

\hspace{4mm}  $\textbf{for} \ t = 0, ..., \ T - 1$ steps \textbf{do} 

\hspace{8mm} Compute $a_{t}$ from behavioral policy, $a_{t} \leftarrow  \pi_{b}(s_{t}, g_{micro})$

\hspace{8mm} Execute $a_{t}$, observe next state $s_{t+1}$ and compute reward $r(s_{t+1}, g_{micro})$

\hspace{8mm} Store transition $(s_{t}, a_{t}, r_{t}, s_{t+1}, g_{micro})$ in $R$   

\hspace{4mm} $\texttt{AchievedGoal}_{n} \leftarrow s_{T}$

\hspace{4mm} \textbf{end for}

\hspace{4mm} Sample a random minibatch of $N$ transitions $(s_{i}, a_{i}, r_{i}, s_{i+1})$ from $R$

\hspace{4mm} Sample new goals $g^{\prime}$ using $S$ 

\hspace{4mm} Replace the sampled transitions goals with the new goal $g^{\prime}$, $(s_{t}, a_{t}, r_{t}, s_{t+1}, g^{\prime})$

\hspace{4mm} Recompute reward for replaced goals

\hspace{4mm} $\textbf{for} \ i = 1, ..., K$ iterations \textbf{do} 

\hspace{8mm} Perform one optimization step for Goal Generator using (\texttt{AchievedGoal}, \texttt{DesiredGoal})

\hspace{4mm} \textbf{end for}

\hspace{4mm} $\textbf{for} \ i = 1, ..., M$ iterations \textbf{do} 

\hspace{8mm} Perform one optimization step of $A$

\hspace{4mm} \textbf{end for}

\noindent \textbf{end for}


\noindent\rule[1.4pt]{0.5\textwidth}{1.0pt}
\vspace{2mm}


Algorithm 1. describes our method in detail. At each iteration, the goal generator produces a \texttt{micro goal} which is used to condition the behavior policy and collect samples by executing it. For each episode, the end of trajectory state, called as the \texttt{achieved goal} is collected and stored in memory. While training, a mini batch of data is sampled and some or all of the goal samples are relabelled with new ones using the goal sampling strategy (described below). The \texttt{achieved goals} and the \texttt{desired goals} are used to update the goal generator periodically. The \texttt{desired goals} essentially either are the goals corresponding to the task in hand or any of the \texttt{sub-goals} provided by the sub-goal extraction method. Therefore, this allows the micro scheme to be run independently as well as in combination with the macro method. We elaborate more on this in the below section.  

\subsection{Strategy for goal sampling}

For replacing goals by sampling new ones, we consider different strategies  such as having a mixture of \texttt{HER goals} and \texttt{micro goals} (referred to as micro-g), and having a mixture of \texttt{HER goals} and \texttt{{desired goals}} (referred to as micro-sg).

\subsection{Environment Details}

\begin{itemize}
    \item \textbf{Pushing} : This requires a block placed on a table to be pushed by the end-effector of the robot to a given target.
    \item \textbf{Sliding} : In this task, the robot is supposed to hit a puck so that the puck reaches a target location. The target location is given at a position out of the reach of the end-effector, hindering the puck from pushing it continuously towards the target. Instead the agent needs learn to solve the task from a single hit. Overall, We observe that although it is possible to learn a good policy, it is very hard to produce a perfect policy. This can possibly be attributed to the design of the task itself, or the fact that using a very small $r_{threshold}$ for such a hard task in calculating the reward.
    \item \textbf{Pick and Place} : This requires the robot agent to pick a box lying on the table and place it at a target location in the air. The gripper is also controlled by the policy in this case, unlike the previous ones. We also do not start any episode with the block already in the robot's gripper, thus making sure that favorable starts are not considered. Specifically for this task, we consider two sampling strategies for the target location. We denote a \texttt{uniform} strategy to sample target location in the air completely randomly without prioritizing the table. A \texttt{non-uniform} strategy is one in which the target is sampled on the table with probability $0.5$ and in the air with probability $0.5$.
\end{itemize}

\subsection{Training Details}

\subsubsection{Goal Generator}
We train a GAN on the achieved and desired goals data gather after each rollout. The generator network consists of two 128 nodes layers, while the discriminator consists of two 256 nodes layers. We use a learning rate of 0.001, a batch size of 64, and sample from a noise vector $z$ of size 4. We run 200 training iterations of the GAN after every 100 iterations of the DDPG policy.

\subsubsection{sub-goal Extractor}
For learning a mapping between start states and sub-goals, we train a  
2 layer MLP with 16 nodes each. The input is the start state while the output is the sub-goal i.e. a vector of size 3. The batch size used is 64, and the learning rate is 0.001. We run 1000 training iterations of this extractor for a dataset consisting of 1000 expert trajectory samples. It is observed that having less number of expert trajectories i.e. around 200 does not affect the accuracy by a lot. 

\subsubsection{Architecture}
We run all experiments till 150 epochs on 5 CPU cores. Each epoch consists of 50 cycles. For each cycle 40 training iterations of DDPG are performed. Both the Actor and Critic networks in DDPG are 3 layer MLPs with ReLU non-linearities, 256 nodes each and learning rate as $10^{-3}$.

\begin{figure*}[h]
	\includegraphics[width=0.33\textwidth]{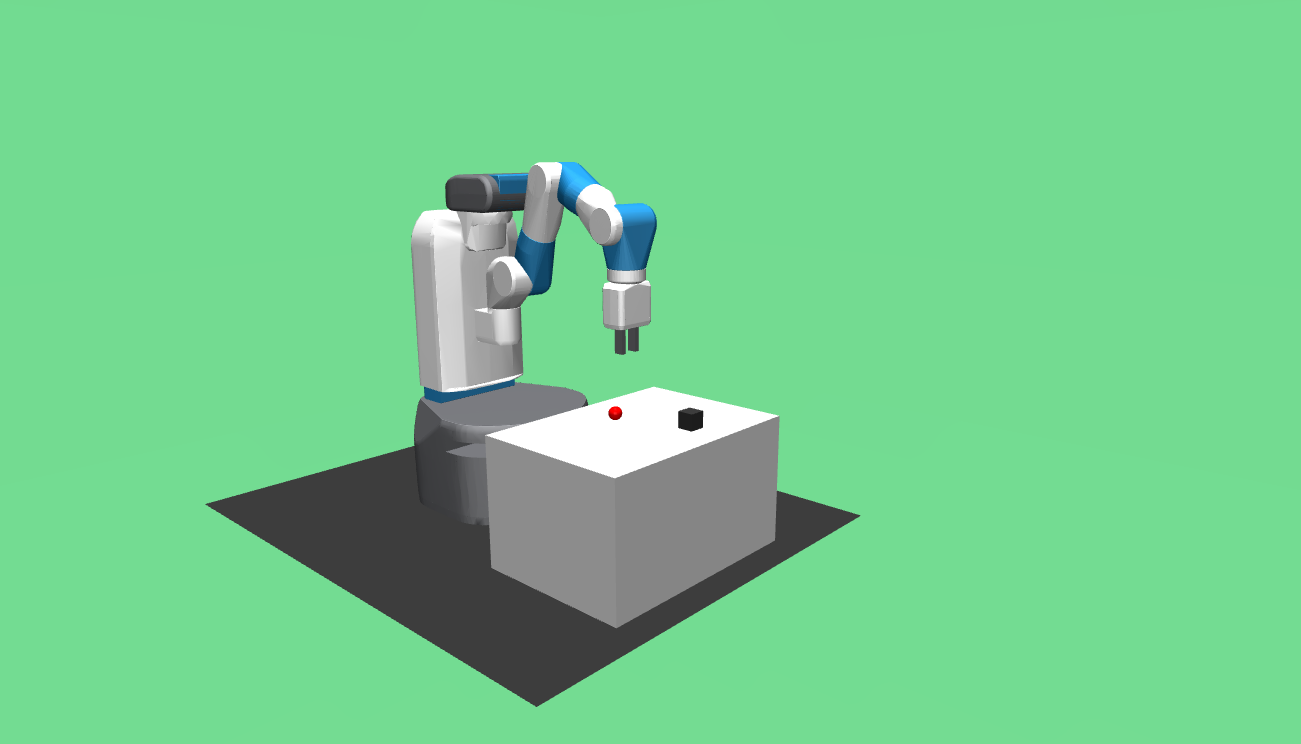}
	\includegraphics[width=0.33\textwidth]{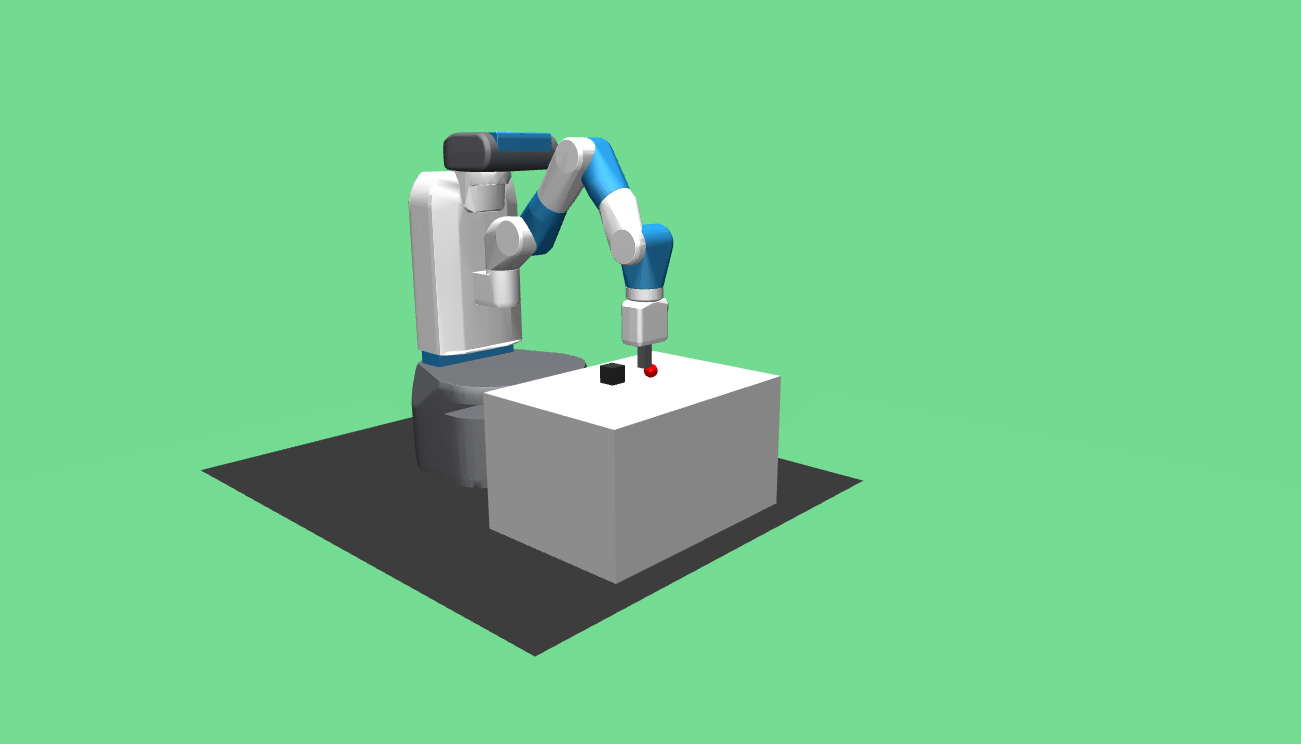}
	\includegraphics[width=0.33\textwidth]{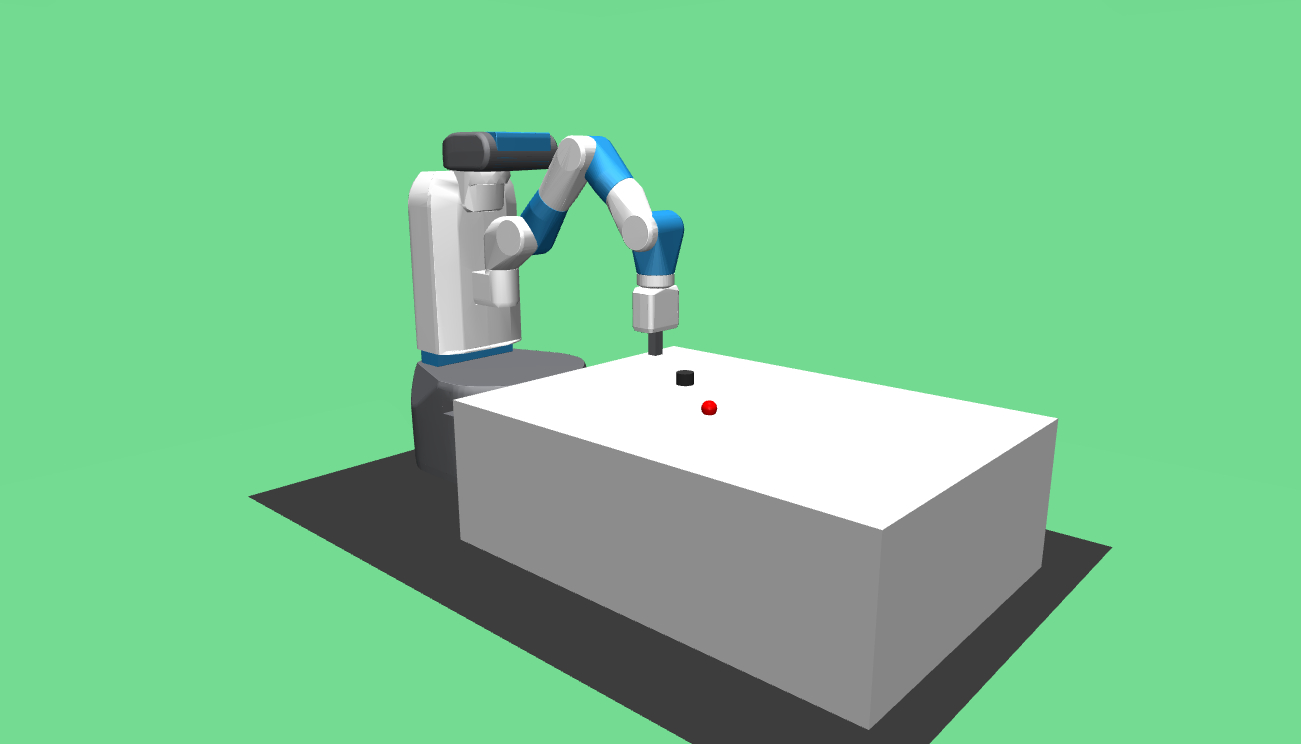}
	\caption{The Fetch Pick and Place, Push and Slide environments}
	\label{envs}
\end{figure*}

\begin{figure*}[h]
	\includegraphics[width=0.33\textwidth]{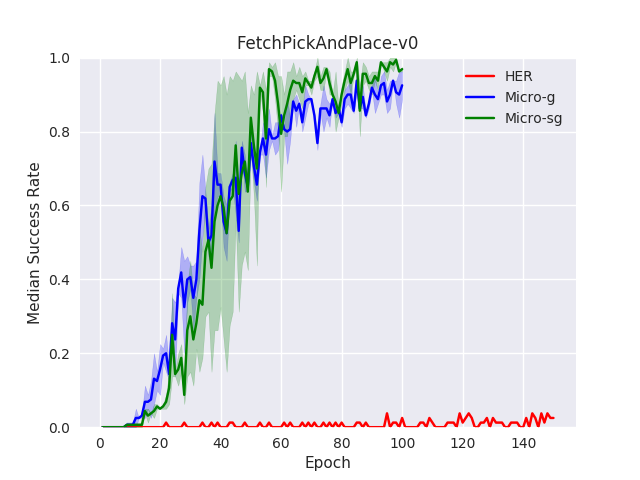}
	\includegraphics[width=0.33\textwidth]{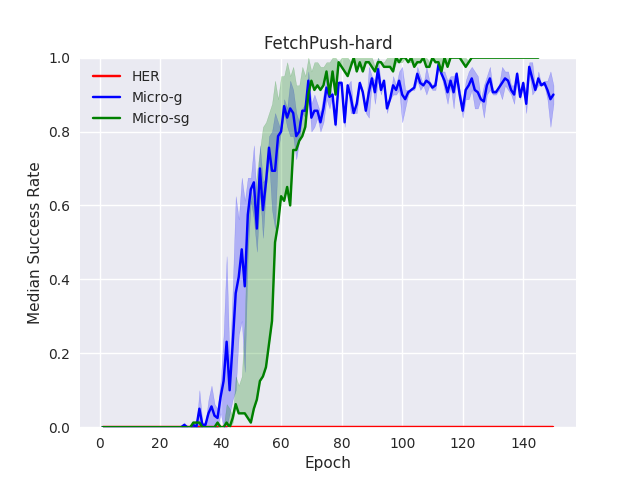}
	\includegraphics[width=0.33\textwidth]{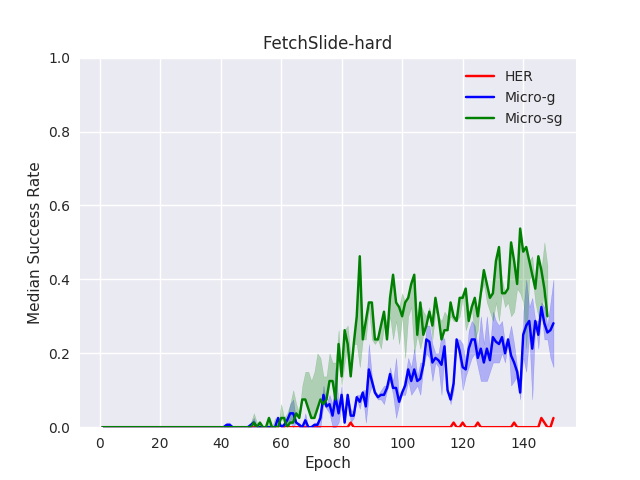}
	\caption{Micro curriculum performance compared with HER for \textbf{a.} (left) Fetch Pick and Place \textbf{b.} (middle) Fetch Slide-hard, \textbf{c.}(right) Fetch Push-hard. Micro-g refers to a goal sampling strategy comprising a mixture of HER samples and \texttt{Micro goals}. Micro-sg refers to a goal sampling strategy comprising a mixture of HER samples and \texttt{{Desired goals}}}
	\label{mainResults}
\end{figure*}

\subsection{Micro - Tasks Considered}

We consider variants of the pushing, sliding and pick and place tasks for a 7 DOF Fetch robot simulation \cite{plappert2018multi} as shown in the Fig \ref{mainResults}. The sampling strategy \textbf{S} for micro used here comprises \texttt{HER goals} and \texttt{micro goal} samples. We consider three tasks in the Mujoco \cite{todorov2012mujoco} environment for our experiments as described below. A successful trajectory receives a $0$ reward while an unsuccessful one receives $-1$ reward. For all three tasks, the target and the object are randomly initialized such that they do not lie in the reward threshold $r_{threshold} = 0.05$, equivalent to 5cm, and therefore the reward received initially is always -1, i.e. we make sure that the agent does not start from a solved state even randomly. We compare our method with the original HER algorithm proposed in \cite{andrychowicz2017hindsight} which is the state-of-the-art algorithm on these domains. Moreover, we also compare with the original DDPG algorithm as a baseline. However, since DDPG fails to solve any of the tasks considered in this paper independently (success rate of almost 0 across all training epochs), we opt to not show these results explicitly in the plots.  

\subsubsection{Push-hard and Slide-hard tasks}
We consider harder variants of the pushing and sliding tasks for testing the micro scheme. These tasks are "made hard" by ensuring that the object and the target do not lie in similar distributions initially and are far apart from each other for all episode samples. This makes the task difficult to solve as even if the agent somehow learns to push or slide the object to some nearby target site, the task is still not considered solved.

\subsubsection{Pick and Place}
The task requires an object to be picked and placed at a target site. The target is never sampled on the table and always in the air. We also do not start any episode with the block already in the robot's gripper, thus making sure that favorable starts are not considered.

\subsection{Results}
We are able to learn optimal policies for all three tasks. For push-hard and slide-hard tasks, HER is unable to even learn to reach the object as shown in Fig \ref{mainResults}. This can be attributed to a mismatch in the kind of goals provided to the parameterized policy and the ones on which the agent learns off-policy. On the other hand, following the micro scheme, we are able to gradually start learning to reach and push / slide the object to nearby generated goals and then gain expertise with respect to the target goals. For Pick and Place, since the goal is always in the air and the object always on the table, a similar mismatch is conceivable. 

\section{Macro Curriculum}
\begin{figure}[h]
\includegraphics[width=0.15\textwidth]{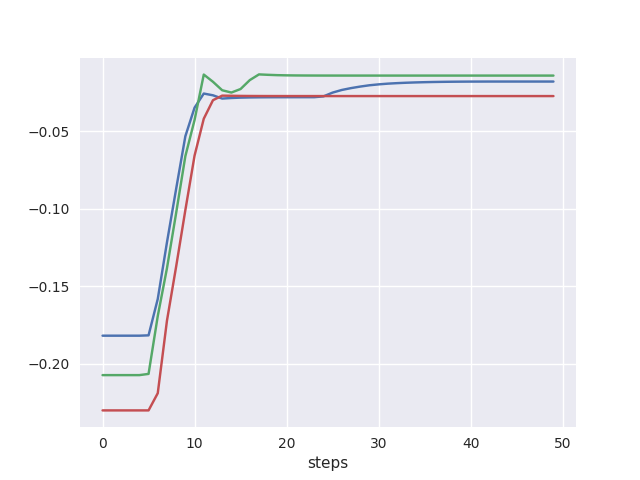}
\includegraphics[width=0.15\textwidth]{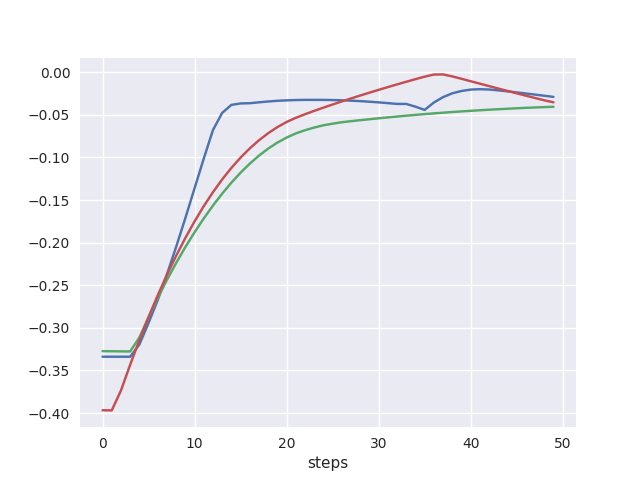}
\includegraphics[width=0.15\textwidth]{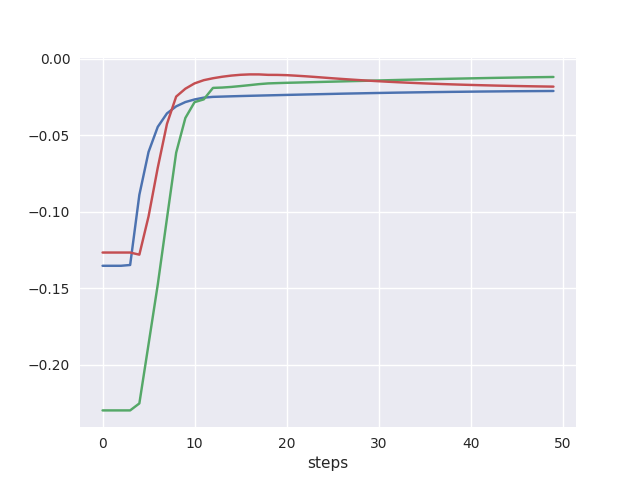}
\includegraphics[width=0.15\textwidth]{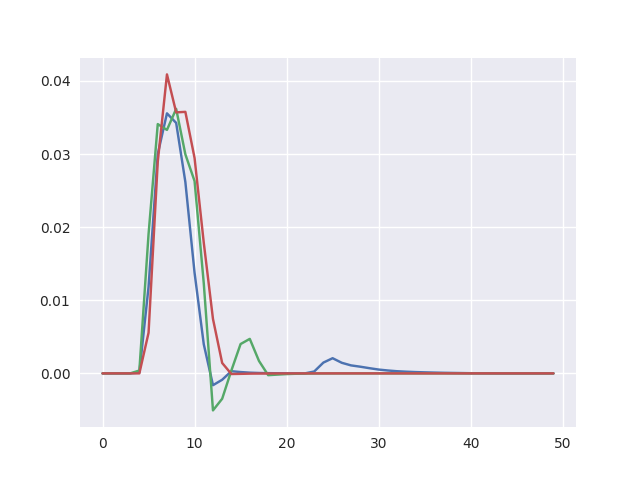}
\includegraphics[width=0.15\textwidth]{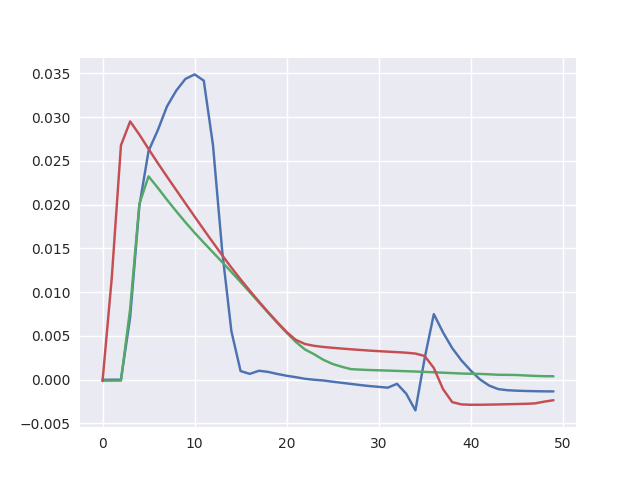}
\includegraphics[width=0.15\textwidth]{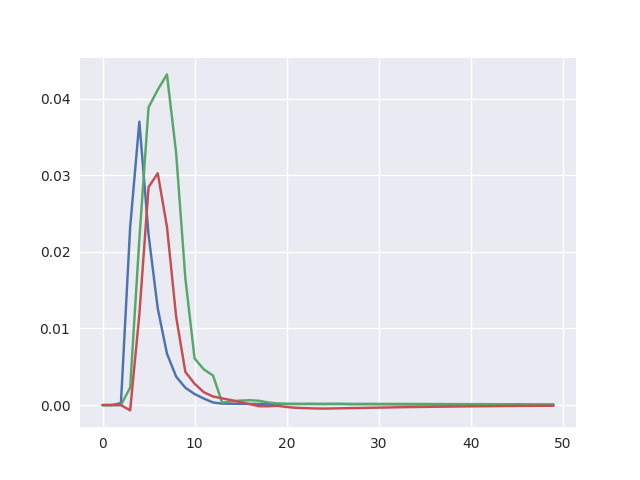}
\includegraphics[width=0.15\textwidth]{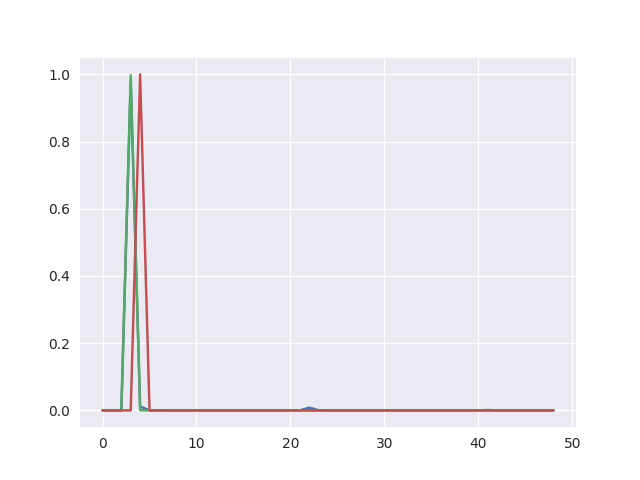}
\includegraphics[width=0.15\textwidth]{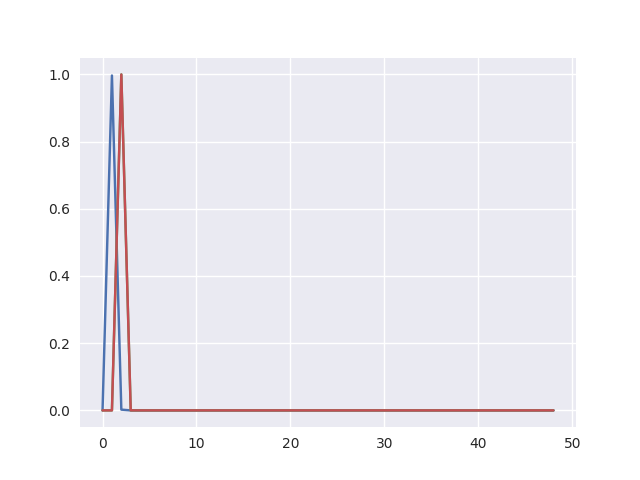}
\includegraphics[width=0.15\textwidth]{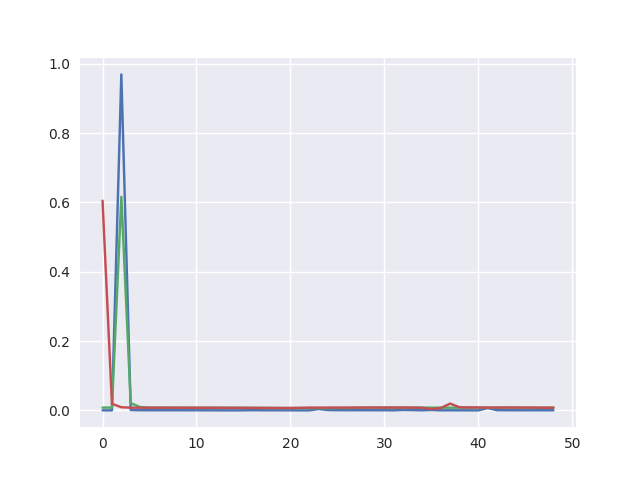}
\caption{sub-goal Extraction for Fetch Push, Slide and Pick and Place for 3 expert trajectory samples \textbf{a. Top row :} dense reward, \textbf{a. Middle row :} reward  gradient, \textbf{a. Bottom row :} ratio of reward gradient. sub-goal for a particular trajectory is taken as the achieved goal at time t, t being obtained by observing the peaks in the ratio of gradient curve. This is done for all such demonstration trajectories and the mapping $F$ is learned over this set of sub-goals and the corresponding start states.}
\label{macro}
\end{figure}

A macro curriculum scheme allows extracting sub-goals by leveraging demonstrated states or observations and sequentially learning the sub-policies for each sub-goal. In the experiments we consider, this implies that learning to achieve the second sub-goal is facilitated by leveraging previous learning of achieving the first sub-goal (learning to push uses already gathered information about learning to reach). We argue that this setting is general enough because each sub-policy itself learns a hard task (the task of reaching) instead of simple "macro" actions (moving the manipulator continuously in a particular direction). This allows representing the final task policy as comprising each sub-policy. Specifically, we consider long horizon tasks and assume that few demonstration state trajectories $\tau = s_{0}, s_{1}, ... s_{t}$ are available for the given tasks. In general, detecting changes in state representation has been shown to be a good method for extracting sub-goals. This is since system dynamics change suddenly around such sub-goals. In our case, the dense reward (eq. 4) computed per time step for a demonstration is used as the signal for sub-goal extraction. We compute the gradient ratio for such a signal and choose the sub-goal as the state for which consistent spikes are observed. Fig \ref{macro} shows the plots for such a dense reward signal in the three tasks considered. The intuition for finding a good sub-goal in a typical manipulation task is to observe that there is a sudden change in the dynamics of the system. For example, if the robot is trying to push a block, it can be easily seen that once the robot explores and starts to interact with the block, the policy will differ as the block interaction dynamics also affect the reward now. For demonstration trajectories, we observe that the gradient ratio of the dense reward always results in consistent spikes near the object position, proving that it is a good sub-goal for learning the three tasks mentioned.

\begin{equation}
r_{dense} = || \ g_{achieved} - g_{desired} \ ||^{2}
\end{equation}

Learning between two such \texttt{sub-goals} can be performed by following a micro curriculum scheme detailed above. The extracted \texttt{sub-goals} form a set of states that are achieved by most of the sampled expert trajectories. Note that these \texttt{sub-goals} are dependent on the start state. This is because we consider learning over varied goals, thus using goal conditioned policies and not over a single goal state. Given a policy $\pi(s_{t}, sg_{t+1})$ that has learnt to achieve a sub-goal $sg_{t}$ allows the agent to achieve the next sub-goal $sg_{t+1}$ by leveraging previous information.

Consider the example of robotic ant navigation where to reach the goal state, the ant needs to collect a key which will open the door to the goal state room. The point we make here is that only using a micro scheme will generate goals between the ants start position and the goal position. However, doing so will result in the ant always jamming against the door with no success in opening it. Since the key lies along another path, through which no micro goals are generated, the agent never learns to open the door. This is where observing an expert and using it to learn that sub-goal lies at the key location becomes relevant. Following this, a micro scheme can be used to learn each sub-policy, that of reaching to the key from the start state and that of reaching to the actual goal state from the key location. 


\vspace{2mm}
\noindent\rule[0.5pt]{0.5\textwidth}{1.0pt}
\vspace{-1.mm}
\noindent \textbf{Method 1 :} Extract sub-goals
\vspace{-1.5mm}

\noindent\rule[1.4pt]{0.5\textwidth}{1.0pt}
\setlength{\parskip}{0.0ex}

Collect state demonstration trajectories $\tau$

Compute dense reward obtained at each stage, $r_{dense} = \newline \  $(\texttt{AchievedGoal}$_{i}$ - \texttt{DesiredGoal}$_{i})^{2} $

Compute ratio of gradient of the dense reward, $r_{grad}$ for each state in an expert trajectory

$p \leftarrow$ Normalize $r_{grad}$ in [0, 1]

\texttt{sub-goals} $\leftarrow$ Sample \texttt{num\_subgoals} states from each trajectory based on highest probability $p$

$\textbf{for} \ n = 1, ..., N$ iterations \ \textbf{do} 

\hspace{2mm}  Train sub-goal extractor $F$ (\texttt{sub-goals}, \texttt{start\_states}) 

\textbf{for end}

\textbf{return} $F$

\vspace{-1mm}
\noindent\rule[0.5pt]{0.5\textwidth}{1.0pt}


\subsection{Macro - Tasks Considered}

\subsubsection{Receptor-PickAndPlace task}
\begin{figure}[h]
	\centering
	\includegraphics[width=0.25\textwidth]{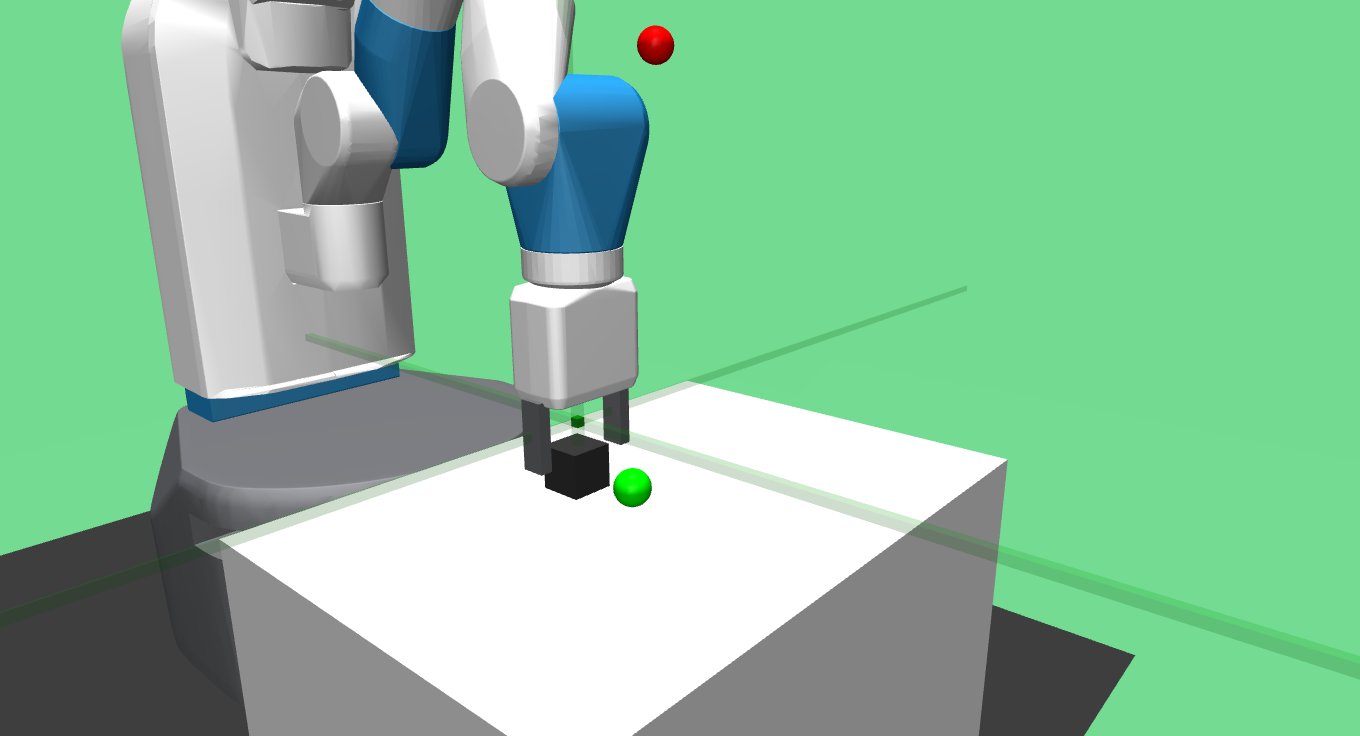}
    \includegraphics[width=0.33\textwidth]{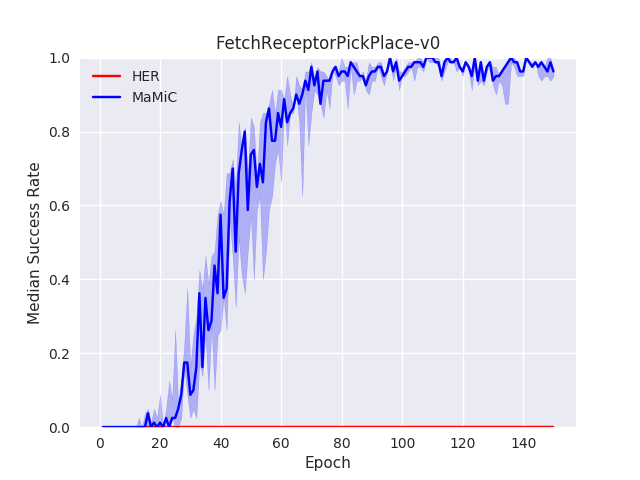}
    \caption{Receptor-PickAndPlace task : }
    \label{receptor}
\end{figure}

We introduce a new task setting called Receptor-PickandPlace which comprises an object placed on a table, a receptor site on the table, and a target located in the air.  As shown in Fig \ref{receptor}, the green and red markers represent the receptor and the goal locations respectively. The agent is required to pick and place the object at a target, which gets activated only if the object passes through the receptor site. Therefore, the agent is not rewarded even if the object is successfully placed at the target, if it does not pass from the receptor site.  Such a task becomes extremely difficult to solve because of a sequencing behavior involved and a sparse reward available. We show how combining the macro and micro schemes can solve this task, by 1) leveraging demonstration states to extract a sub-goal near the receptor site and 2) using a powerful micro scheme to realize the sequencing of tasks involved, i.e. first moving the block to the receptor and then to the target.
\vspace{2mm}

\begin{equation}
  r = \left\{
    \begin{array}{l}
       $0, \quad if receptor on and distance to goal $<$ $r_{threshold}$$\\
      $-1,\quad otherwise$
    \end{array}
  \right.
\end{equation}

\subsubsection{Push-far and Slide-far tasks}
We also consider variants of the pushing and sliding tasks in which the start state (the gripper position) is considerably far from the table and varied as opposed to the default case where the gripper always starts from a single state and over the table. 

\subsection{Results}
For the Receptor-PickandPlace task, recognizing the receptor as a sub-goal is crucial to learning. There is a significant peak in the dense reward gradient ratio around the receptor location, proving that the sub-goal extraction in the macro scheme is able to leverage demonstrations efficiently. 
This when combined with a micro scheme is able to learn the sequence of going to the receptor first with the block, thus activating the target, followed by placing it over the target. HER and micro scheme applied individually would fail to learn this task as shown in Fig \ref{receptor}, for different reasons. For HER, the task is too difficult because of the target being quite far and always sampled in the air. With a micro scheme alone, initially we see that the policy learned tries to pick and place the object to targets which are just above the table directly without going over the receptor. However, since the agent is not being rewarded, it quickly diverges to random behavior.

\begin{figure}[h]
	\centering
	\includegraphics[width=0.33\textwidth]{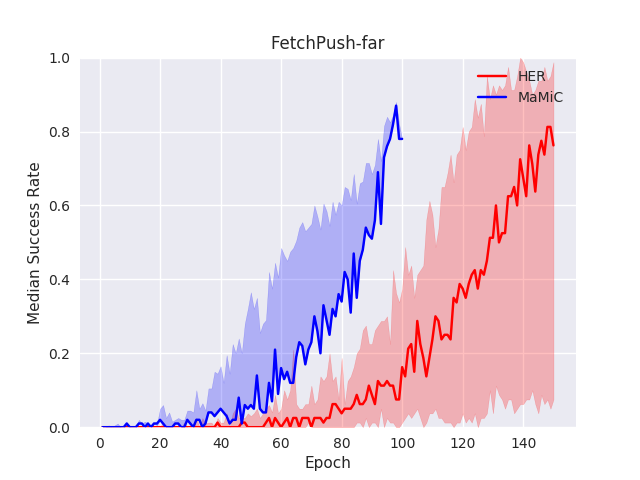}
	\includegraphics[width=0.33\textwidth]{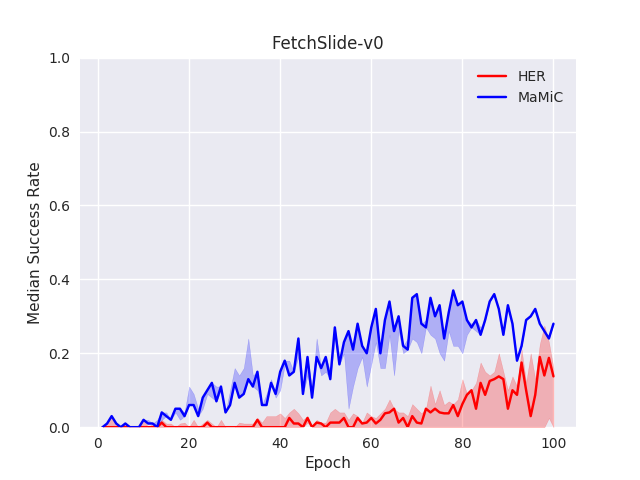}
	\caption{MaMiC's performance compared with HER when the agent starts from an widened initial distribution for \textbf{a.} (left) Fetch Push-far \textbf{b.} (right) Fetch Slide-far}
	\label{far}
\end{figure}

For the Push-far and Slide-far tasks, both MaMiC and HER learn useful policies. Since in these tasks, object and target lie in overlapping distributions, HER is able to perform well as shown in Fig \ref{far}. However, please note the extremely high variance in HER, ranging from solving the task in some instances to learning no useful behavior at all in some. This can potentially be attributed to the fact that since the gripper starts at a significantly different part of the state space as the block, learning no longer remains as stable as when the gripper starts over the table and close to the block. MaMiC, on the other hand is able to first learn the reaching sub-task by identifying locations close to the object's position as good \texttt{sub-goals} and then learns to push the object. Please note that MaMiC provides a clear acceleration in this case and is more stable than HER.

\section {Related and Future Work}

\cite{konidaris2009skill} exploit the idea of starting from states near the goal and then gradually expanding the starting distribution to learn the overall task. This works because the agent slowly starts to learn how to reach states which are close to the goal. \cite{florensa2017reverse} build on this concept and propose a scheme for expanding the start state distribution based on the reward received while starting from such states. However, as mentioned, the usual assumption here is that the agent has the ability to reset to any state, which is not general enough. Moreover, the experiments are shown on tasks having single goal states, and therefore the policy is not generalized for a multiple goal domain such as pick and place. It is not at all trivial to extend this idea for goal parameterized policies as well. \cite{sukhbaatar2017intrinsic} also propose an automatic curriculum generation scheme, but work on the assumption that the environment is either reversible or resettable. There have been other works such as \cite{mcgovern2001automatic}, \cite{csimcsek2005identifying} which propose different methods for extracting sub-goals. On a higher level, given a sub-goal extraction technique and a function which maps goals to states, our method can work on domains other than robotic manipulation as well. A by-product of an evolving policy, as in our method, is that the sub policies can be saved as learnt options (\cite{sutton1999between}, \cite{bacon2017option}) and then used for transfer to tasks which define a different meaning but require similar options. Similar ideas have been reported in \cite{florensa2017stochastic}, \cite{da2012learning}, where the agent learns a set of skills in a pre training procedure. Such skills are later combined with a master policy which allows for efficient exploration. These works mainly build on a bottom-up approach which restricts the meta-policy required to solve complex tasks to comprise only pre-defined or pre-learnt options.

Since the setting of the algorithm is quite general, there are multiple directions for extending this work. The next challenge is to show how such a technique performs on even more longer horizon tasks, perhaps involving multiple objects as well. Working with image based observations can allow for learning richer representations useful in sub-goal extraction. Moreover, collecting state or observation demonstration trajectories is relatively simpler and more intuitive with images. Considering better heuristics for how $\alpha$ is updated to produce goals closer to the \texttt{DesiredGoal} distribution is an important point to improve upon. Another avenue for future work is to incorporate different schemes of sub-goal extraction which exploit domain specific properties.

\section{Conclusion}
\label{sec:conclusion}

We introduce a dual curriculum scheme for robotic manipulation which aids in exploration in robotic manipulation tasks with very sparse rewards. We show how the micro scheme is a powerful method for generating goals intelligently and can allow solving hard variants of the pushing, sliding and pick and place tasks without resetting to arbitrary states, starting from favorable states or using expert actions. Moreover, through the Receptor-PickandPlace task, we emphasize on the need for a macro scheme combined with micro when a task involves completing sub-tasks sequentially. 


\bibliographystyle{ACM-Reference-Format}  
\bibliography{sample-bibliography}  


\begin{thebibliography}{00}


\ifx \showCODEN    \undefined \def \showCODEN     #1{\unskip}     \fi
\ifx \showDOI      \undefined \def \showDOI       #1{#1}\fi
\ifx \showISBNx    \undefined \def \showISBNx     #1{\unskip}     \fi
\ifx \showISBNxiii \undefined \def \showISBNxiii  #1{\unskip}     \fi
\ifx \showISSN     \undefined \def \showISSN      #1{\unskip}     \fi
\ifx \showLCCN     \undefined \def \showLCCN      #1{\unskip}     \fi
\ifx \shownote     \undefined \def \shownote      #1{#1}          \fi
\ifx \showarticletitle \undefined \def \showarticletitle #1{#1}   \fi
\ifx \showURL      \undefined \def \showURL       {\relax}        \fi
\providecommand\bibfield[2]{#2}
\providecommand\bibinfo[2]{#2}
\providecommand\natexlab[1]{#1}
\providecommand\showeprint[2][]{arXiv:#2}

\bibitem[\protect\citeauthoryear{Andrychowicz, Wolski, Ray, Schneider, Fong,
  Welinder, McGrew, Tobin, Abbeel, and Zaremba}{Andrychowicz
  et~al\mbox{.}}{2017}]%
        {andrychowicz2017hindsight}
\bibfield{author}{\bibinfo{person}{Marcin Andrychowicz}, \bibinfo{person}{Filip
  Wolski}, \bibinfo{person}{Alex Ray}, \bibinfo{person}{Jonas Schneider},
  \bibinfo{person}{Rachel Fong}, \bibinfo{person}{Peter Welinder},
  \bibinfo{person}{Bob McGrew}, \bibinfo{person}{Josh Tobin},
  \bibinfo{person}{OpenAI~Pieter Abbeel}, {and} \bibinfo{person}{Wojciech
  Zaremba}.} \bibinfo{year}{2017}\natexlab{}.
\newblock \showarticletitle{Hindsight experience replay}. In
  \bibinfo{booktitle}{{\em Advances in Neural Information Processing Systems}}.
  \bibinfo{pages}{5048--5058}.
\newblock


\bibitem[\protect\citeauthoryear{Bacon, Harb, and Precup}{Bacon
  et~al\mbox{.}}{2017}]%
        {bacon2017option}
\bibfield{author}{\bibinfo{person}{Pierre-Luc Bacon}, \bibinfo{person}{Jean
  Harb}, {and} \bibinfo{person}{Doina Precup}.}
  \bibinfo{year}{2017}\natexlab{}.
\newblock \showarticletitle{The Option-Critic Architecture.}. In
  \bibinfo{booktitle}{{\em AAAI}}. \bibinfo{pages}{1726--1734}.
\newblock


\bibitem[\protect\citeauthoryear{Bengio, Louradour, Collobert, and
  Weston}{Bengio et~al\mbox{.}}{2009}]%
        {bengio2009curriculum}
\bibfield{author}{\bibinfo{person}{Yoshua Bengio},
  \bibinfo{person}{J{\'e}r{\^o}me Louradour}, \bibinfo{person}{Ronan
  Collobert}, {and} \bibinfo{person}{Jason Weston}.}
  \bibinfo{year}{2009}\natexlab{}.
\newblock \showarticletitle{Curriculum learning}. In \bibinfo{booktitle}{{\em
  Proceedings of the 26th annual international conference on machine
  learning}}. ACM, \bibinfo{pages}{41--48}.
\newblock


\bibitem[\protect\citeauthoryear{Da~Silva, Konidaris, and Barto}{Da~Silva
  et~al\mbox{.}}{2012}]%
        {da2012learning}
\bibfield{author}{\bibinfo{person}{Bruno Da~Silva}, \bibinfo{person}{George
  Konidaris}, {and} \bibinfo{person}{Andrew Barto}.}
  \bibinfo{year}{2012}\natexlab{}.
\newblock \showarticletitle{Learning parameterized skills}.
\newblock \bibinfo{journal}{{\em arXiv preprint arXiv:1206.6398\/}}
  (\bibinfo{year}{2012}).
\newblock


\bibitem[\protect\citeauthoryear{Deisenroth, Neumann, Peters,
  et~al\mbox{.}}{Deisenroth et~al\mbox{.}}{2013}]%
        {deisenroth2013survey}
\bibfield{author}{\bibinfo{person}{Marc~Peter Deisenroth},
  \bibinfo{person}{Gerhard Neumann}, \bibinfo{person}{Jan Peters},
  {et~al\mbox{.}}} \bibinfo{year}{2013}\natexlab{}.
\newblock \showarticletitle{A survey on policy search for robotics}.
\newblock \bibinfo{journal}{{\em Foundations and Trends{\textregistered} in
  Robotics\/}} \bibinfo{volume}{2}, \bibinfo{number}{1--2}
  (\bibinfo{year}{2013}), \bibinfo{pages}{1--142}.
\newblock


\bibitem[\protect\citeauthoryear{Florensa, Duan, and Abbeel}{Florensa
  et~al\mbox{.}}{2017a}]%
        {florensa2017stochastic}
\bibfield{author}{\bibinfo{person}{Carlos Florensa}, \bibinfo{person}{Yan
  Duan}, {and} \bibinfo{person}{Pieter Abbeel}.}
  \bibinfo{year}{2017}\natexlab{a}.
\newblock \showarticletitle{Stochastic neural networks for hierarchical
  reinforcement learning}.
\newblock \bibinfo{journal}{{\em arXiv preprint arXiv:1704.03012\/}}
  (\bibinfo{year}{2017}).
\newblock


\bibitem[\protect\citeauthoryear{Florensa, Held, Wulfmeier, and
  Abbeel}{Florensa et~al\mbox{.}}{2017b}]%
        {florensa2017reverse}
\bibfield{author}{\bibinfo{person}{Carlos Florensa}, \bibinfo{person}{David
  Held}, \bibinfo{person}{Markus Wulfmeier}, {and} \bibinfo{person}{Pieter
  Abbeel}.} \bibinfo{year}{2017}\natexlab{b}.
\newblock \showarticletitle{Reverse curriculum generation for reinforcement
  learning}.
\newblock \bibinfo{journal}{{\em arXiv preprint arXiv:1707.05300\/}}
  (\bibinfo{year}{2017}).
\newblock


\bibitem[\protect\citeauthoryear{Goodfellow, Pouget-Abadie, Mirza, Xu,
  Warde-Farley, Ozair, Courville, and Bengio}{Goodfellow et~al\mbox{.}}{2014}]%
        {goodfellow2014generative}
\bibfield{author}{\bibinfo{person}{Ian Goodfellow}, \bibinfo{person}{Jean
  Pouget-Abadie}, \bibinfo{person}{Mehdi Mirza}, \bibinfo{person}{Bing Xu},
  \bibinfo{person}{David Warde-Farley}, \bibinfo{person}{Sherjil Ozair},
  \bibinfo{person}{Aaron Courville}, {and} \bibinfo{person}{Yoshua Bengio}.}
  \bibinfo{year}{2014}\natexlab{}.
\newblock \showarticletitle{Generative adversarial nets}. In
  \bibinfo{booktitle}{{\em Advances in neural information processing systems}}.
  \bibinfo{pages}{2672--2680}.
\newblock


\bibitem[\protect\citeauthoryear{Gu, Holly, Lillicrap, and Levine}{Gu
  et~al\mbox{.}}{2017}]%
        {gu2017deep}
\bibfield{author}{\bibinfo{person}{Shixiang Gu}, \bibinfo{person}{Ethan Holly},
  \bibinfo{person}{Timothy Lillicrap}, {and} \bibinfo{person}{Sergey Levine}.}
  \bibinfo{year}{2017}\natexlab{}.
\newblock \showarticletitle{Deep reinforcement learning for robotic
  manipulation with asynchronous off-policy updates}. In
  \bibinfo{booktitle}{{\em Robotics and Automation (ICRA), 2017 IEEE
  International Conference on}}. IEEE, \bibinfo{pages}{3389--3396}.
\newblock


\bibitem[\protect\citeauthoryear{Held, Geng, Florensa, and Abbeel}{Held
  et~al\mbox{.}}{2017}]%
        {held2017automatic}
\bibfield{author}{\bibinfo{person}{David Held}, \bibinfo{person}{Xinyang Geng},
  \bibinfo{person}{Carlos Florensa}, {and} \bibinfo{person}{Pieter Abbeel}.}
  \bibinfo{year}{2017}\natexlab{}.
\newblock \showarticletitle{Automatic goal generation for reinforcement
  learning agents}.
\newblock \bibinfo{journal}{{\em arXiv preprint arXiv:1705.06366\/}}
  (\bibinfo{year}{2017}).
\newblock


\bibitem[\protect\citeauthoryear{Konidaris and Barto}{Konidaris and
  Barto}{2009}]%
        {konidaris2009skill}
\bibfield{author}{\bibinfo{person}{George Konidaris} {and}
  \bibinfo{person}{Andrew~G Barto}.} \bibinfo{year}{2009}\natexlab{}.
\newblock \showarticletitle{Skill discovery in continuous reinforcement
  learning domains using skill chaining}. In \bibinfo{booktitle}{{\em Advances
  in neural information processing systems}}. \bibinfo{pages}{1015--1023}.
\newblock


\bibitem[\protect\citeauthoryear{Lillicrap, Hunt, Pritzel, Heess, Erez, Tassa,
  Silver, and Wierstra}{Lillicrap et~al\mbox{.}}{2016}]%
        {lillicrap2016continuous}
\bibfield{author}{\bibinfo{person}{Timothy Lillicrap},
  \bibinfo{person}{Jonathan Hunt}, \bibinfo{person}{Alexander Pritzel},
  \bibinfo{person}{Nicolas Heess}, \bibinfo{person}{Tom Erez},
  \bibinfo{person}{Yuval Tassa}, \bibinfo{person}{David Silver}, {and}
  \bibinfo{person}{Daan Wierstra}.} \bibinfo{year}{2016}\natexlab{}.
\newblock \showarticletitle{Continuous control with deep reinforcement
  learning.(2016)}.
\newblock \bibinfo{journal}{{\em arXiv preprint cs.LG/1509.02971\/}}
  (\bibinfo{year}{2016}).
\newblock


\bibitem[\protect\citeauthoryear{Liu, Gupta, Abbeel, and Levine}{Liu
  et~al\mbox{.}}{2018}]%
        {liu2018imitation}
\bibfield{author}{\bibinfo{person}{YuXuan Liu}, \bibinfo{person}{Abhishek
  Gupta}, \bibinfo{person}{Pieter Abbeel}, {and} \bibinfo{person}{Sergey
  Levine}.} \bibinfo{year}{2018}\natexlab{}.
\newblock \showarticletitle{Imitation from observation: Learning to imitate
  behaviors from raw video via context translation}. In
  \bibinfo{booktitle}{{\em 2018 IEEE International Conference on Robotics and
  Automation (ICRA)}}. IEEE, \bibinfo{pages}{1118--1125}.
\newblock


\bibitem[\protect\citeauthoryear{Mao, Li, Xie, Lau, Wang, and Smolley}{Mao
  et~al\mbox{.}}{2017}]%
        {mao2017least}
\bibfield{author}{\bibinfo{person}{Xudong Mao}, \bibinfo{person}{Qing Li},
  \bibinfo{person}{Haoran Xie}, \bibinfo{person}{Raymond~YK Lau},
  \bibinfo{person}{Zhen Wang}, {and} \bibinfo{person}{Stephen~Paul Smolley}.}
  \bibinfo{year}{2017}\natexlab{}.
\newblock \showarticletitle{Least squares generative adversarial networks}. In
  \bibinfo{booktitle}{{\em 2017 IEEE International Conference on Computer
  Vision (ICCV)}}. IEEE, \bibinfo{pages}{2813--2821}.
\newblock


\bibitem[\protect\citeauthoryear{McGovern and Barto}{McGovern and
  Barto}{2001}]%
        {mcgovern2001automatic}
\bibfield{author}{\bibinfo{person}{Amy McGovern} {and}
  \bibinfo{person}{Andrew~G Barto}.} \bibinfo{year}{2001}\natexlab{}.
\newblock \showarticletitle{Automatic discovery of subgoals in reinforcement
  learning using diverse density}. In \bibinfo{booktitle}{{\em ICML}},
  Vol.~\bibinfo{volume}{1}. \bibinfo{pages}{361--368}.
\newblock


\bibitem[\protect\citeauthoryear{Mnih, Kavukcuoglu, Silver, Rusu, Veness,
  Bellemare, Graves, Riedmiller, Fidjeland, Ostrovski, et~al\mbox{.}}{Mnih
  et~al\mbox{.}}{2015}]%
        {mnih2015human}
\bibfield{author}{\bibinfo{person}{Volodymyr Mnih}, \bibinfo{person}{Koray
  Kavukcuoglu}, \bibinfo{person}{David Silver}, \bibinfo{person}{Andrei~A
  Rusu}, \bibinfo{person}{Joel Veness}, \bibinfo{person}{Marc~G Bellemare},
  \bibinfo{person}{Alex Graves}, \bibinfo{person}{Martin Riedmiller},
  \bibinfo{person}{Andreas~K Fidjeland}, \bibinfo{person}{Georg Ostrovski},
  {et~al\mbox{.}}} \bibinfo{year}{2015}\natexlab{}.
\newblock \showarticletitle{Human-level control through deep reinforcement
  learning}.
\newblock \bibinfo{journal}{{\em Nature\/}} \bibinfo{volume}{518},
  \bibinfo{number}{7540} (\bibinfo{year}{2015}), \bibinfo{pages}{529}.
\newblock


\bibitem[\protect\citeauthoryear{Nair, McGrew, Andrychowicz, Zaremba, and
  Abbeel}{Nair et~al\mbox{.}}{2017}]%
        {nair2017overcoming}
\bibfield{author}{\bibinfo{person}{Ashvin Nair}, \bibinfo{person}{Bob McGrew},
  \bibinfo{person}{Marcin Andrychowicz}, \bibinfo{person}{Wojciech Zaremba},
  {and} \bibinfo{person}{Pieter Abbeel}.} \bibinfo{year}{2017}\natexlab{}.
\newblock \showarticletitle{Overcoming exploration in reinforcement learning
  with demonstrations}.
\newblock \bibinfo{journal}{{\em arXiv preprint arXiv:1709.10089\/}}
  (\bibinfo{year}{2017}).
\newblock


\bibitem[\protect\citeauthoryear{Plappert, Andrychowicz, Ray, McGrew, Baker,
  Powell, Schneider, Tobin, Chociej, Welinder, et~al\mbox{.}}{Plappert
  et~al\mbox{.}}{2018}]%
        {plappert2018multi}
\bibfield{author}{\bibinfo{person}{Matthias Plappert}, \bibinfo{person}{Marcin
  Andrychowicz}, \bibinfo{person}{Alex Ray}, \bibinfo{person}{Bob McGrew},
  \bibinfo{person}{Bowen Baker}, \bibinfo{person}{Glenn Powell},
  \bibinfo{person}{Jonas Schneider}, \bibinfo{person}{Josh Tobin},
  \bibinfo{person}{Maciek Chociej}, \bibinfo{person}{Peter Welinder},
  {et~al\mbox{.}}} \bibinfo{year}{2018}\natexlab{}.
\newblock \showarticletitle{Multi-Goal Reinforcement Learning: Challenging
  Robotics Environments and Request for Research}.
\newblock \bibinfo{journal}{{\em arXiv preprint arXiv:1802.09464\/}}
  (\bibinfo{year}{2018}).
\newblock


\bibitem[\protect\citeauthoryear{Schaul, Horgan, Gregor, and Silver}{Schaul
  et~al\mbox{.}}{2015}]%
        {schaul2015universal}
\bibfield{author}{\bibinfo{person}{Tom Schaul}, \bibinfo{person}{Daniel
  Horgan}, \bibinfo{person}{Karol Gregor}, {and} \bibinfo{person}{David
  Silver}.} \bibinfo{year}{2015}\natexlab{}.
\newblock \showarticletitle{Universal value function approximators}. In
  \bibinfo{booktitle}{{\em International Conference on Machine Learning}}.
  \bibinfo{pages}{1312--1320}.
\newblock


\bibitem[\protect\citeauthoryear{Schulman, Levine, Abbeel, Jordan, and
  Moritz}{Schulman et~al\mbox{.}}{2015}]%
        {schulman2015trust}
\bibfield{author}{\bibinfo{person}{John Schulman}, \bibinfo{person}{Sergey
  Levine}, \bibinfo{person}{Pieter Abbeel}, \bibinfo{person}{Michael Jordan},
  {and} \bibinfo{person}{Philipp Moritz}.} \bibinfo{year}{2015}\natexlab{}.
\newblock \showarticletitle{Trust region policy optimization}. In
  \bibinfo{booktitle}{{\em International Conference on Machine Learning}}.
  \bibinfo{pages}{1889--1897}.
\newblock


\bibitem[\protect\citeauthoryear{Silver, Huang, Maddison, Guez, Sifre, Van
  Den~Driessche, Schrittwieser, Antonoglou, Panneershelvam, Lanctot,
  et~al\mbox{.}}{Silver et~al\mbox{.}}{2016}]%
        {silver2016mastering}
\bibfield{author}{\bibinfo{person}{David Silver}, \bibinfo{person}{Aja Huang},
  \bibinfo{person}{Chris~J Maddison}, \bibinfo{person}{Arthur Guez},
  \bibinfo{person}{Laurent Sifre}, \bibinfo{person}{George Van Den~Driessche},
  \bibinfo{person}{Julian Schrittwieser}, \bibinfo{person}{Ioannis Antonoglou},
  \bibinfo{person}{Veda Panneershelvam}, \bibinfo{person}{Marc Lanctot},
  {et~al\mbox{.}}} \bibinfo{year}{2016}\natexlab{}.
\newblock \showarticletitle{Mastering the game of Go with deep neural networks
  and tree search}.
\newblock \bibinfo{journal}{{\em nature\/}} \bibinfo{volume}{529},
  \bibinfo{number}{7587} (\bibinfo{year}{2016}), \bibinfo{pages}{484--489}.
\newblock


\bibitem[\protect\citeauthoryear{{\c{S}}im{\c{s}}ek, Wolfe, and
  Barto}{{\c{S}}im{\c{s}}ek et~al\mbox{.}}{2005}]%
        {csimcsek2005identifying}
\bibfield{author}{\bibinfo{person}{{\"O}zg{\"u}r {\c{S}}im{\c{s}}ek},
  \bibinfo{person}{Alicia~P Wolfe}, {and} \bibinfo{person}{Andrew~G Barto}.}
  \bibinfo{year}{2005}\natexlab{}.
\newblock \showarticletitle{Identifying useful subgoals in reinforcement
  learning by local graph partitioning}. In \bibinfo{booktitle}{{\em
  Proceedings of the 22nd international conference on Machine learning}}. ACM,
  \bibinfo{pages}{816--823}.
\newblock


\bibitem[\protect\citeauthoryear{Sukhbaatar, Lin, Kostrikov, Synnaeve, Szlam,
  and Fergus}{Sukhbaatar et~al\mbox{.}}{2017}]%
        {sukhbaatar2017intrinsic}
\bibfield{author}{\bibinfo{person}{Sainbayar Sukhbaatar},
  \bibinfo{person}{Zeming Lin}, \bibinfo{person}{Ilya Kostrikov},
  \bibinfo{person}{Gabriel Synnaeve}, \bibinfo{person}{Arthur Szlam}, {and}
  \bibinfo{person}{Rob Fergus}.} \bibinfo{year}{2017}\natexlab{}.
\newblock \showarticletitle{Intrinsic motivation and automatic curricula via
  asymmetric self-play}.
\newblock \bibinfo{journal}{{\em arXiv preprint arXiv:1703.05407\/}}
  (\bibinfo{year}{2017}).
\newblock


\bibitem[\protect\citeauthoryear{Sutton and Barto}{Sutton and Barto}{1998}]%
        {sutton1998reinforcement}
\bibfield{author}{\bibinfo{person}{Richard~S Sutton} {and}
  \bibinfo{person}{Andrew~G Barto}.} \bibinfo{year}{1998}\natexlab{}.
\newblock \bibinfo{booktitle}{{\em Reinforcement learning: An introduction}}.
  Vol.~\bibinfo{volume}{1}.
\newblock \bibinfo{publisher}{MIT press Cambridge}.
\newblock


\bibitem[\protect\citeauthoryear{Sutton, Precup, and Singh}{Sutton
  et~al\mbox{.}}{1999}]%
        {sutton1999between}
\bibfield{author}{\bibinfo{person}{Richard~S Sutton}, \bibinfo{person}{Doina
  Precup}, {and} \bibinfo{person}{Satinder Singh}.}
  \bibinfo{year}{1999}\natexlab{}.
\newblock \showarticletitle{Between MDPs and semi-MDPs: A framework for
  temporal abstraction in reinforcement learning}.
\newblock \bibinfo{journal}{{\em Artificial intelligence\/}}
  \bibinfo{volume}{112}, \bibinfo{number}{1-2} (\bibinfo{year}{1999}),
  \bibinfo{pages}{181--211}.
\newblock


\bibitem[\protect\citeauthoryear{Todorov, Erez, and Tassa}{Todorov
  et~al\mbox{.}}{2012}]%
        {todorov2012mujoco}
\bibfield{author}{\bibinfo{person}{Emanuel Todorov}, \bibinfo{person}{Tom
  Erez}, {and} \bibinfo{person}{Yuval Tassa}.} \bibinfo{year}{2012}\natexlab{}.
\newblock \showarticletitle{Mujoco: A physics engine for model-based control}.
  In \bibinfo{booktitle}{{\em Intelligent Robots and Systems (IROS), 2012
  IEEE/RSJ International Conference on}}. IEEE, \bibinfo{pages}{5026--5033}.
\newblock


\end{thebibliography}

\end{document}